\documentclass[letterpaper, 10 pt, conference]{ieeeconf}  %
\usepackage{float}
\usepackage{graphicx}
\usepackage{booktabs}
\usepackage{cuted}   %
\usepackage{capt-of} %
\usepackage{comment}
\usepackage{subcaption}
\usepackage{xcolor} 
\usepackage{amsmath}
\usepackage{amssymb}   %
\usepackage{censor}
\StopCensoring

\makeatletter
\let\NAT@parse\undefined
\makeatother

\usepackage[colorlinks=true,linkcolor=blue,citecolor=blue,urlcolor=blue]{hyperref}

\makeatletter
\def\@citex[#1]#2{%
  \let\@citea\@empty
  \@cite{\@for\@citeb:=#2\do
    {\@citea\def\@citea{], [}%
     \edef\@citeb{\expandafter\@firstofone\@citeb\@empty}%
     \if@filesw\immediate\write\@auxout{\string\citation{\@citeb}}\fi
     \@ifundefined{b@\@citeb}{\mbox{\reset@font\bfseries ?}%
       \G@refundefinedtrue
       \@latex@warning{Citation `\@citeb' on page \thepage \space undefined}}%
       {\csname b@\@citeb\endcsname}}}{#1}}
\makeatother

\IEEEoverridecommandlockouts                              %

\title{\LARGE \bf
MWM: Mobile World Models for Action-Conditioned Consistent Prediction
}

\author{
    \textbf{Han Yan}$^{*}$\quad
    \textbf{Zishang Xiang}$^{*}$\quad 
    \textbf{Zeyu Zhang}$^{*\dag}$\quad 
    \textbf{Hao Tang}$^{\ddag}$ \vspace{0.1cm}\\
    School of Computer Science, Peking University\\
    \small $^*$Equal contribution. $^\dag$Project lead.
    $^\ddag$Corresponding author: bjdxtanghao@gmail.com.
}

\begin{document}
\maketitle

\begin{strip}
  \centering
  \begin{minipage}{\textwidth}
    \centering
    \includegraphics[width=\textwidth,trim=0mm 6mm 0mm 0mm,clip]{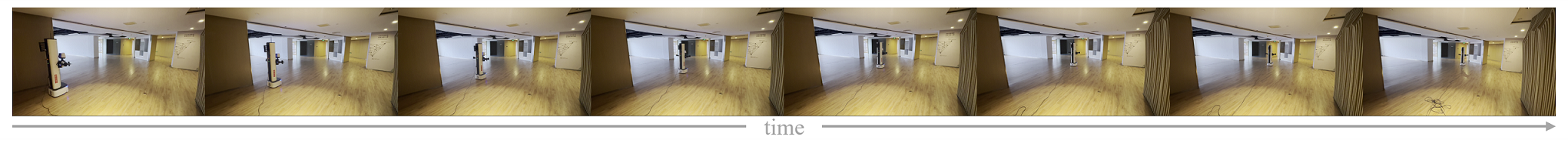}
    
    \par\vspace{0.1mm}
    {\raggedright \small (a) The goal is set to the cabinet.\par}
  \end{minipage}

  \vspace{2mm}

  \begin{minipage}{\textwidth}
    \centering
    \includegraphics[width=\textwidth,trim=0mm 6mm 0mm 0mm,clip]{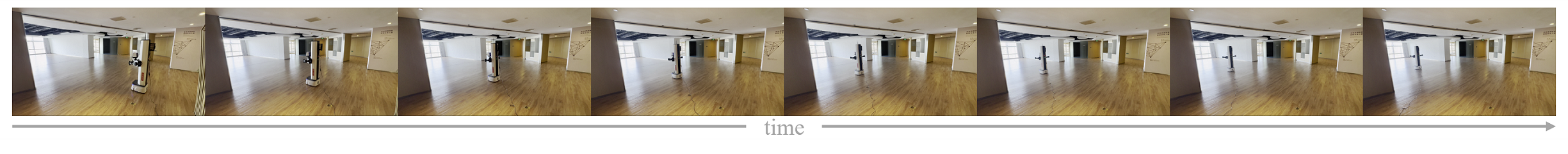}
    
    \par\vspace{0.1mm}
    {\raggedright \small (b) The goal is set to the window.\par}
  \end{minipage}

  \vspace{2mm}

  \begin{minipage}{\textwidth}
    \centering
    \includegraphics[width=\textwidth,trim=0mm 6mm 0mm 0mm,clip]{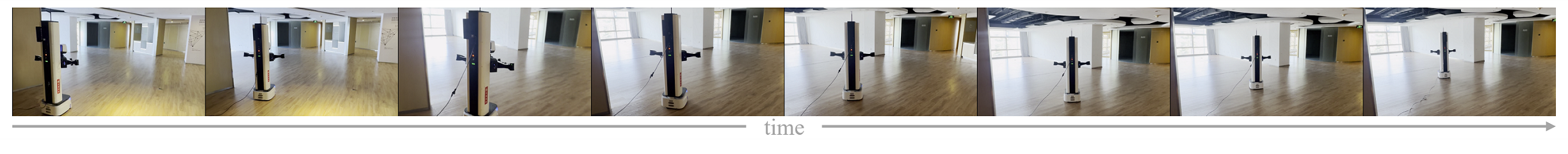}
    
    \par\vspace{0.1mm}
    {\raggedright \small (c) The same goal is set to the window, but from a different starting point.\par}
  \end{minipage}

  \vspace{1mm}
  \captionof{figure}{\textbf{Real-world demonstration of MWM.} Upon receiving current observations, MWM imagines action-conditioned future trajectories. The planning is performed over candidate rollouts to identify the optimal navigation plan, enabling successful image-goal navigation in real-world environments.}
  \label{fig:real}
\end{strip}

\thispagestyle{empty}
\pagestyle{empty}

\begin{abstract}

World models enable planning in imagined future predicted space, offering a promising framework for embodied navigation. However, existing navigation world models often lack action-conditioned consistency, so visually plausible predictions can still drift under multi-step rollout and degrade planning. Moreover, efficient deployment requires few-step diffusion inference, but existing distillation methods do not explicitly preserve rollout consistency, creating a training–inference mismatch. To address these challenges, we propose MWM, a mobile world model for planning-based image-goal navigation. Specifically, we introduce a two-stage training framework that combines structure pretraining with Action-Conditioned Consistency (ACC) post-training to improve action-conditioned rollout consistency. We further introduce Inference-Consistent State Distillation (ICSD) for few-step diffusion distillation with improved rollout consistency. Our experiments on benchmark and real-world tasks demonstrate consistent gains in visual fidelity, trajectory accuracy, planning success, and inference efficiency.
Code: \url{https://github.com/AIGeeksGroup/MWM}.
Website: \url{https://aigeeksgroup.github.io/MWM}.

\end{abstract}

\section{INTRODUCTION}
World models have emerged as a promising approach for Embodied-AI \cite{DreamerV3,Cosmos,UniPi,DIAMOND}. For robot navigation, they enable planning in the predicted future observation space instead of relying solely on end-to-end action policies. Given the current observation and candidate actions, a world model can generate future visual rollouts and support planning by scoring candidate trajectories against task-specific objectives, such as goal observations or language-conditioned task requirements. 

Although existing embodied navigation world models can produce visually faithful future observations \cite{Irasim,SuSIE,GR-1,Worlddreamer}, their \emph{action-conditioned consistency} remains insufficient for planning. In particular, the predicted visual rollout may look realistic frame by frame, yet still deviate from the true trajectory that would be induced by the same action sequence in the real world. This mismatch is especially problematic for model predictive control (MPC) \cite{MPC}, which relies on searching over imagined trajectories to select the best action sequence. When the model prediction is not well aligned with the real execution outcome, MPC may favor trajectories that appear correct in the generated rollout but lead to erroneous behavior when executed on the robot. Second, real-world deployment requires fast inference for responsive planning and control, which makes it necessary to perform a few-step distillation. However, existing diffusion distillation methods are mostly designed to match the teacher model at the distribution level \cite{LCM,ADD,TCD,Hyper-SD,PCM}. While this improves sampling efficiency, such distribution-level alignment does not explicitly preserve \emph{action-conditioned consistency} during rollout, and may therefore weaken the consistency required for reliable planning.

For the first challenge, our motivation is to explicitly reduce the discrepancy between model-predicted rollouts and real observations, which we interpret as a form of \emph{error accumulation} \cite{ScheduledSampling,Areduction}. Even over relatively short planning action sequences, small deviations at each prediction step can compound over time, causing the final predicted state to drift substantially from the true spatial position. Therefore, we need a training strategy that can preserve high-fidelity image generation while reducing error accumulation during action-conditioned rollout. For the second challenge, our motivation is that for MPC-based world models, few-step distillation should be performed at the level of \emph{consistency} rather than merely at the level of distribution. Since downstream planning depends on the reliability of rollouts rather than distributional similarity alone, the distilled model should preserve rollout consistency with respect to the real trajectory.

To this end, we propose MWM, a mobile world model designed to enhance action-conditioned consistency in visual planning. Our first contribution is a two-stage training pipeline consisting of \emph{structure pretraining} followed by Action-Conditioned Consistency (ACC) post-training. In the first stage, the model learns stable scene structure and appearance through supervised training. In the second stage, ACC post-training explicitly trains the model under self-conditioned action rollout contexts, with the goal of mitigating error accumulation so that autoregressive predictions remain better aligned with the real observations. Our second contribution is Inference-Consistent State Distillation (ICSD), which extends the same ACC post-training objective—namely, a \emph{consistency distillation} objective—to efficient few-step diffusion distillation. By introducing an Inference-Consistent State, ICSD better preserves action-conditioned rollout consistency during planning. Our third contribution is a comprehensive evaluation on both benchmark and real-world navigation tasks. Our method reduces DreamSim \cite{DreamSim} by 20.4\% and FID \cite{FID} by 17.5\% while delivering at least a 4$\times$ inference speedup, improves trajectory accuracy by 10.9\% in ATE \cite{ATE} and 8.5\% in RPE \cite{ATE} on benchmark evaluation, and achieves a 50\% relative improvement in success rate and a 32.1\% reduction in navigation error in real-world deployment.

The main contributions of our work can be summarized as follows:
\begin{itemize}
    \item We propose a two-stage training pipeline for MWM, consisting of \emph{structure pretraining} and Action-Conditioned Consistency (ACC) post-training. The post-training explicitly trains the model under self-conditioned action rollout contexts to mitigate error accumulation while preserving high-fidelity visual generation and improving alignment between autoregressive predictions and real observations.
    
    \item We introduce Inference-Consistent State Distillation (ICSD) within the post-training stage, which supports efficient few-step diffusion inference under the consistency distillation objective by reducing the discrepancy between training-time diffusion states and the denoising states encountered during accelerated deployment, thereby improving both rollout reliability and inference efficiency.
    
    \item We conduct comprehensive evaluations on both benchmark and real-world navigation tasks. Our method reduces DreamSim by 20.4\% and FID by 17.5\% while achieving at least a 4$\times$ faster inference speed, improves trajectory accuracy by 10.9\% in ATE and 8.5\% in RPE on benchmark evaluation, and achieves a 50\% relative improvement in success rate together with a 32.1\% reduction in navigation error in real-world deployment.
\end{itemize}

\section{Related Work}
\noindent
\textbf{Learning-Based Visual Navigation.}
Goal-conditioned visual navigation has been extensively studied in robotics, with early methods relying on hand-crafted features and metric maps for motion planning~\cite{huang2025mobilevla,liu2025nav}. The advent of large-scale deep learning brought substantial advances through end-to-end policies trained on diverse robot datasets. GNM demonstrated platform-agnostic generalization via goal-conditioned topological planning \cite{GNM}, while ViNT extended this with transformer-based pretraining for robust zero-shot transfer across indoor and outdoor environments \cite{ViNT}. NoMaD unified goal-directed navigation and open-ended exploration with a goal-masked diffusion policy, improving robustness to multimodal action distributions \cite{NoMaD}. More recent work integrates vision-language models to execute tasks from open-vocabulary natural language instructions without explicit waypoint supervision \cite{Lm-nav}. Despite this progress, these reactive, end-to-end approaches share a fundamental limitation: they do not explicitly model the future visual consequences of candidate actions, making them ill-suited for environments where lookahead planning is necessary.

\noindent
\textbf{World Model-Based Embodied Navigation.}
World models have a rich history in model-based reinforcement learning \cite{worldmodels,zhang2026geoworld,zhang2026code2worlds}, where they simulate state transitions to support planning and policy optimization. The Dreamer series showed that compact latent world models enable efficient policy learning across diverse continuous control tasks \cite{DreamtoControl,DreamerV3}. Recent work extends this paradigm to real-world visual navigation: Navigation World Models (NWM) \cite{NWM}proposed a navigation world model based on a conditional diffusion transformer (CDiT) that generates egocentric future observations conditioned on robot actions, supporting model predictive control entirely in the imagined visual space. UniSim further demonstrated that large-scale video prediction models trained on diverse interaction data can serve as interactive simulators for robot tasks \cite{LearningInteractive}. However, a critical limitation persists across these models: visual realism does not imply action-conditioned consistency. Predicted frame sequences may appear locally plausible yet deviate substantially from the trajectory that the same action sequence would induce in reality—a mismatch that directly undermines MPC, whose planning quality depends on whether imagined rollouts faithfully reflect real execution outcomes.

\noindent
\textbf{Diffusion Acceleration and Autoregressive Consistency.}
Diffusion probabilistic models have become the dominant paradigm for high-fidelity visual generation and have been widely adopted in robot learning for policy generation and world modeling \cite{DenoisingDiffusionProbabilistic}. A key practical bottleneck for deploying diffusion-based world models in online planning is inference latency: standard DDPM sampling requires hundreds of sequential denoising steps, making real-time MPC infeasible on robot hardware. DDIM \cite{DenoisingDiffusionImplicit}, DPM-Solver \cite{Dpm-solver}, and consistency distillation \cite{ConsistencyModels} progressively reduce function evaluations to tens or even single steps, but these methods are designed to preserve distributional fidelity at the individual frame level and do not explicitly maintain action-conditioned rollout consistency under accelerated sampling. This limitation is closely connected to the problem of autoregressive error accumulation. Under teacher-forced training, the model always conditions on ground-truth past frames, whereas at test time it must condition on its own predictions; the resulting distribution shift causes per-step deviations to compound over multi-step rollouts. Scheduled sampling \cite{ScheduledSampling} and DAgger \cite{Areduction} addressed analogous train-test discrepancies in sequence models and imitation learning, respectively. More recently, Self-Forcing applied this idea to diffusion-based video generation, demonstrating that training under self-generated context substantially reduces long-horizon error accumulation \cite{Selfforcing}. In robotic world models, this issue is especially consequential, since planning accuracy depends directly on whether the predicted terminal state corresponds to the robot's true physical location after executing the planned action sequence.

\begin{figure*}[t]
  \centering
  \begin{subfigure}[t]{0.49\textwidth}
    \centering
    \includegraphics[width=\linewidth,
      trim=45mm 3mm 35mm 5mm,clip]{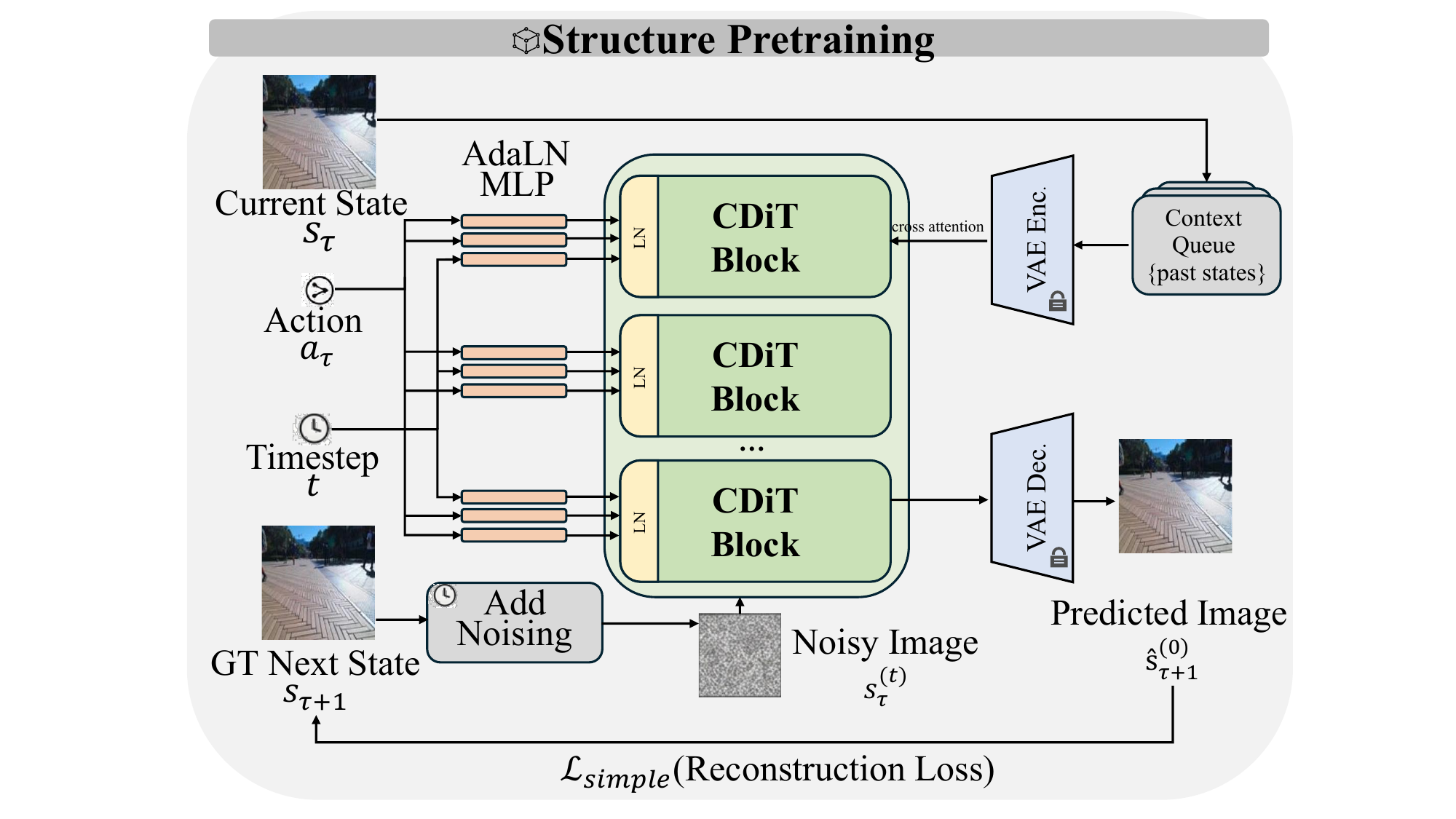}
    \caption{Stage I.}
    \label{fig:stage1}
  \end{subfigure}\hspace{0mm}
  \begin{subfigure}[t]{0.49\textwidth}
    \centering
    \includegraphics[width=\linewidth,,
      trim=38mm 6.6mm 45mm 4mm,clip]{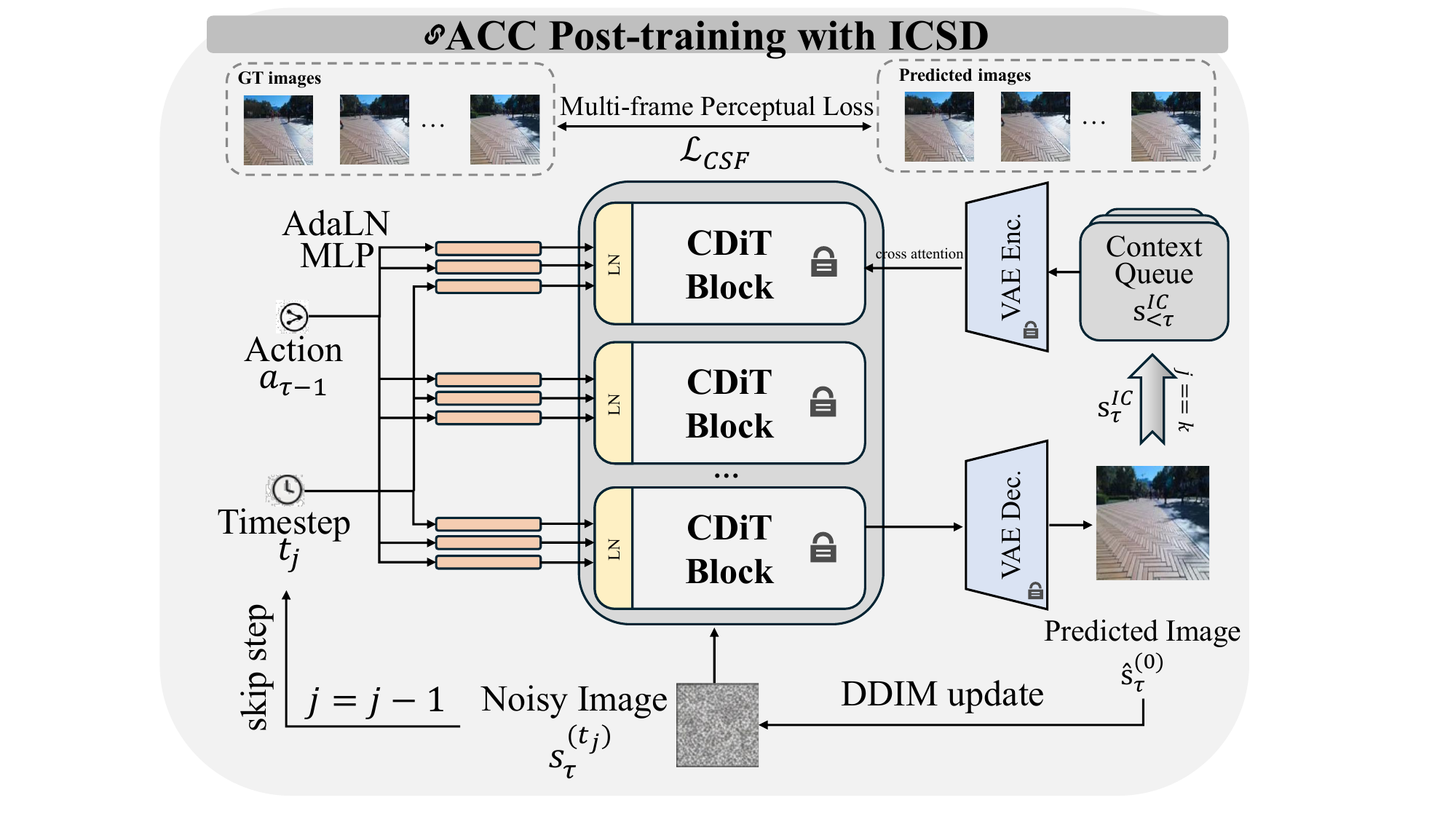}
    \caption{Stage II.}
    \label{fig:stage2}
  \end{subfigure}
  \caption{\textbf{Overview of the Two-stage training pipeline for MWM.} Our training paradigm first performs structure pretraining to learn fine-grained geometry and illumination-dependent appearance, then applies ACC post-training to mitigate compounding error while freezing the CDiT backbone and updating only AdaLN. Within post-training, we introduce ICSD to enable distillation that preserves the consistency objective, while aligning truncated training-time estimates with the inference-time endpoint.}
  \label{fig:twostage}
\end{figure*}

\section{The Proposed Method}
\subsection{Overview} 

We propose a two-stage training pipeline that follows a “structure-first, consistency-refine” paradigm for MWM. In the first stage, we perform structure pretraining to learn high-fidelity scene dynamics and capture fine-grained environmental structure, including detailed geometry and illumination-dependent appearance. In the second stage, we post-train the model on the same dataset under the Action-Conditioned Consistency (ACC) post-training paradigm, which refines autoregressive rollout behavior without destroying the detailed structures acquired from pretraining. This stage is designed to mitigate error accumulation and improve action-conditioned observation alignment with ground-truth trajectories, while simultaneously reducing the train--test gap induced by pretraining. In addition, we design an Inference-Consistent State Distillation (ICSD) mechanism within the ACC post-training by modifying how the diffusion timestep is injected into the backbone, enabling few-step diffusion distillation. Crucially, ICSD introduces an inference-consistent state that aligns the truncated estimate state of training-time (often too smooth/blurred when obtained from an intermediate denoising step) with the inference-time state produced at s=0, which is substantially closer to the result of full progressive denoising, thus reducing the truncation-induced train--test discrepancy (See Fig.~\ref{fig:twostage} for the two-stage training pipeline).  For planning, we adopt the CEM-based search method in the world-model rollout space used in prior work \cite{NWM}.

\subsection{Two-Stage Training Pipeline For MWM}
\subsubsection{Stage I: Structure Pretraining}
We find this stage to be necessary for learning fine-grained scene structure and illumination-dependent appearance in the new environment, which provides a strong initialization for the subsequent ACC post-training. We pretrain MWM as an action-conditioned diffusion model under a teacher-forcing setup, where the ground-truth next representation $s_{\tau+1}$ is provided and the model learns to denoise it given $(s_\tau, a_\tau)$. We sample a diffusion step $t \in \{1,\dots,T\}$ and form a noised target:
\begin{equation}
s^{(t)}_{\tau+1} = \sqrt{\bar{\alpha}_t}\, s_{\tau+1} + \sqrt{1-\bar{\alpha}_t}\,\epsilon,
\quad \epsilon \sim \mathcal{N}(0,I),
\end{equation}
where $\{\bar{\alpha}_t\}$ is the cumulative noise schedule \cite{DenoisingDiffusionProbabilistic} (we follow CDiT settings \cite{NWM}). The denoiser $F_\theta$ takes $(s^{(t)}_{\tau+1}, s_\tau, a_\tau, t)$ as input and predicts a clean target:
\begin{equation}
\hat{s}_{\tau+1} = F_\theta\!\left(s^{(t)}_{\tau+1}\mid s_\tau, a_\tau, t\right).
\end{equation}
We optimize the reconstruction loss over randomly sampled timesteps:
\begin{equation}
\mathcal{L}_{\mathrm{simple}}
=
\mathbb{E}_{(s_\tau,a_\tau,s_{\tau+1}),\,\epsilon,\,t}
\left[\left\| s_{\tau+1} - \hat{s}_{\tau+1} \right\|_2^2\right].
\end{equation}

As shown in Fig.~\ref{fig:twostage}, the action $a_{\tau}$ and timestep $t$ are encoded with sin--cos features and injected into every CDiT block via AdaLN. The context $s_{\tau}$ attends to a memory of past states $\{s_{\tau-1},\ldots,s_{\tau-m}\}$ through cross-attention.

\subsubsection{Stage II: Action-Conditioned Consistency (ACC) Post-training}
We continue training on the same dataset as Stage~I, but expose the model to its own predictions as context, thereby reducing the train--test mismatch between (i) pretraining, where the context is always ground truth, and (ii) deployment, where the context quickly becomes model-generated after a few rollout steps.

Our procedure follows the core self-forcing idea \cite{Selfforcing}: for a rollout of length $N$, we generate frames autoregressively with a diffusion model, and randomly choose a truncation denoising step $k \sim \mathrm{Uniform}(\{1,\dots,T'\})$, where $\{t_1, t_2, \ldots, t_{T'}\}$ denotes a subset of denoising timesteps selected from the full schedule $\{1,\ldots,T\}$. For each frame $\tau$, we start from Gaussian noise and run the reverse process from $t=t_{T'}$ down to $t=t_k$. Importantly, we treat the intermediate denoising steps as \emph{stop-gradient} updates, and only enable gradient computation when producing the final prediction at the truncation step,
\begin{equation}
\hat{s}^{(0)}_{\tau} = G_{\theta}\!\left(s^{(t_k)}_{\tau};\, t_k,\, \hat{s}^{(0)}_{<\tau},\,a_{\tau-1}\right),
\end{equation}
where $\hat{s}^{(0)}_{<\tau}$ denotes the previously generated frames used as autoregressive context. This setup explicitly trains the model under the same self-conditioned context distribution encountered at test time.

Different from distillation-oriented objectives, we update parameters using an observation-level consistency loss. Specifically, we supervise the rollout against ground-truth observations $\{s_{\tau}\}_{\tau=1}^{N}$ with a multi-frame averaged perceptual loss based on LPIPS \cite{LPIPS}:
\begin{equation}
\mathcal{L}_{\mathrm{CSF}}
=
\frac{1}{N}\sum_{\tau=1}^{N}
\mathrm{LPIPS}\!\left(\hat{s}^{(0)}_{\tau},\, s_\tau\right).
\end{equation}
To preserve the high-fidelity image quality learned in Stage~I, we freeze the CDiT backbone and optimize only the lightweight AdaLN modulation layers (i.e., the linear projections that inject the action $a_{\tau}$ conditioning into each block). In practice, this targeted post-training improves multi-step observation alignment without degrading the fine-grained structure and appearance captured by pretraining.

We interpret the discrepancy between self-rolled predictions and real observations as \emph{compounding error}: as rollout length increases, the deviation typically grows, which directly hurts planning accuracy when the terminal predicted frame is used to assess goal-reaching. By reducing this self-conditioning gap, ACC post-training yields more reliable terminal predictions, leading to more accurate trajectory evaluation and improved navigation performance.

\begin{figure*}[!t]
  \centering
  \begin{minipage}[t]{0.49\textwidth}
    \centering
    \includegraphics[width=\linewidth,trim=0mm 2mm 2mm 0mm,clip]{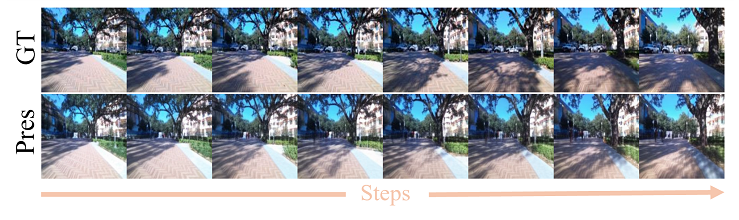}
  \end{minipage}\hfill
  \begin{minipage}[t]{0.49\textwidth}
    \centering
    \includegraphics[width=\linewidth,trim=0mm 2mm 2mm 0mm,clip]{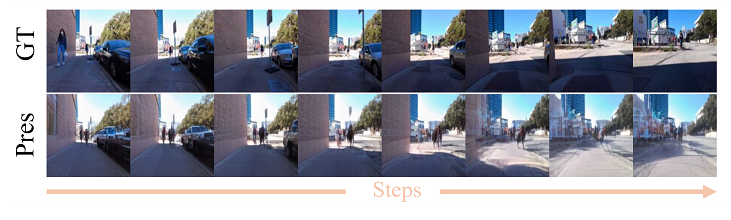}
  \end{minipage}
  \caption{Qualitative results on SCAND. The predicted frames exhibit action-conditioned consistency (ACC) with the ground-truth frames.}
  \label{fig:benchmark_vis}
\end{figure*}

\subsection{Inference-Consistent State Distillation (ICSD)}

In addition, we design an Inference-Consistent State Distillation (ICSD) mechanism within the ACC post-training stage to enable low-step diffusion self-distillation. Rather than introducing a separate training stage or a new objective, ICSD operates under the same consistency supervision as ACC by introducing an \emph{inference-consistent state}. This mechanism enables distillation at the level of \emph{action-conditioned consistency}, rather than merely matching output distributions. More importantly, ICSD introduces an inference-consistent state. Without the inference-consistent state, the ACC supervision can be severely weakened by the over-smoothed intermediate states caused by truncation and skip-step denoising.

Specifically, we inject the diffusion timestep together with the action condition through the same AdaLN modulation network. Consequently, freezing the backbone and optimizing only the AdaLN modulation in Stage~II not only adapts the action-conditioning pathway, but also implicitly updates how timestep information is incorporated into the model. This updated \emph{timestep} conditioning facilitates \emph{few-step diffusion} at inference time.

As described in Stage~II, we randomly choose a truncation denoising step
$k \sim \mathrm{Uniform}(\{1,\ldots,T\})$ and use the corresponding
estimate/state (e.g., $\hat{s}^{(k)\rightarrow(0)}_{\tau}$) for rollouts.
However, the estimate $\hat{s}^{(k)\rightarrow(0)}_{\tau}$ obtained by \emph{truncating} the reverse process at step $k$ is often overly smooth/blurred, \emph{not only} because it is extracted from an intermediate truncation level, \emph{but also} because it is generated by denoising along a \emph{skip-step} subsampled timestep set $\{t_1,\ldots,t_{T'}\}\subseteq\{1,\ldots,T\}$ rather than the full progressive chain, and therefore exhibits a mismatch with the inference-time endpoint. To reduce this gap, ICSD introduces an \emph{inference-consistent state} $s^{\mathrm{IC}}_{\tau}$ to explicitly bridge the discrepancy between the truncated training-time state and the inference-time denoising endpoint. We use a deterministic DDIM update \cite{DenoisingDiffusionImplicit} (with $\eta=0$) from $t_j$ to $t_{j\!-\!1}$:
\begin{equation}
s_{\tau}^{(t_{j-1})}
=
\sqrt{\bar{\alpha}_{t_{j-1}}}\;\hat{s}_{\tau}^{(0)}
+
\sqrt{1-\bar{\alpha}_{t_{j-1}}}\;
\frac{
s_{\tau}^{(t_j)}-\sqrt{\bar{\alpha}_{t_j}}\;\hat{s}_{\tau}^{(0)}
}{
\sqrt{1-\bar{\alpha}_{t_j}}
}.
\end{equation}
where $\hat{s}_{\tau}^{(0)}$ is obtained by updating Eq.~(4) as follows:
\begin{equation}
\hat{s}^{(0)}_{\tau} = G_{\theta}\!\left(s^{(t_k)}_{\tau};\, t_k,\, s^{IC}_{<\tau},\,a_{\tau-1}\right),
\end{equation}
when $t = t_k$, the $s^{\mathrm{IC}}_{\tau}$ is obtained from Eq.~(7).
This explicit alignment reduces the truncation-induced train--test mismatch and improves multi-step rollout consistency under few-step sampling.

\subsection{Planning with MWM}
Following NWM, we formulate navigation as MPC and optimize action sequences with CEM in the world-model rollout space \cite{NWM}. In general, a trajectory evaluator can be instantiated by a VLM to score language instruction alignment or detect obstacles; in this work, we use a terminal-frame perceptual objective:
\begin{equation}
S(s_T, x^*) \;=\; -\,\mathrm{LPIPS}\!\left(\mathrm{Dec}(s_T),\, x^{*}\right),
\end{equation}
where $s_T$ is the rolled-out terminal latent, $\mathrm{Dec}(\cdot)$ decodes it to pixels, and $x^*$ is the goal image. CEM iteratively samples action sequences, ranks them by $S(\cdot)$, and updates the action distribution using the top-$k$ elites until convergence or a fixed iteration budget.

\section{Experiments}

\begin{table*}[t]
\centering
\caption{ACC measured by LPIPS and DreamSim at different rollout horizons on SCAND.}
\label{tab:lpips_dreamsim_rollout}
\resizebox{\linewidth}{!}{
\begin{tabular}{l|cc cc cc cc cc}
\toprule
Model & \multicolumn{2}{c}{1s} & \multicolumn{2}{c}{2s} & \multicolumn{2}{c}{4s} & \multicolumn{2}{c}{8s} & \multicolumn{2}{c}{16s} \\
\cmidrule(lr){2-3}\cmidrule(lr){4-5}\cmidrule(lr){6-7}\cmidrule(lr){8-9}\cmidrule(lr){10-11}
& LPIPS$\downarrow$ & DreamSim$\downarrow$ & LPIPS$\downarrow$ & DreamSim$\downarrow$ & LPIPS$\downarrow$ & DreamSim$\downarrow$ & LPIPS$\downarrow$ & DreamSim$\downarrow$ & LPIPS$\downarrow$ & DreamSim$\downarrow$ \\
\midrule
NWM (DDIM 25) \cite{NWM} & 0.478 & 0.309 & 0.494 & 0.308 & 0.508 & 0.321 & 0.540 & 0.345 & 0.569 & 0.373 \\
NWM (DDIM 5) \cite{NWM} & 0.581 & 0.436 & 0.591 & 0.435 & 0.614 & 0.455 & 0.678 & 0.513 & 0.734 & 0.568 \\
MWM (ours, DDIM 5) & \textbf{0.368} & \textbf{0.244} & \textbf{0.395} & \textbf{0.260} & \textbf{0.421} & \textbf{0.276} & \textbf{0.459} & \textbf{0.306} & \textbf{0.495} & \textbf{0.337} \\
\bottomrule
\end{tabular}
}
\end{table*}

\subsection{Experimental Settings}
\noindent\textbf{Datasets}. For the public dataset, we use \textit{SCAND}, a large-scale dataset of human-teleoperated, socially compliant navigation demonstrations collected in diverse indoor and outdoor environments around a university. SCAND contains 8.7 hours of data from 138 trajectories (25 miles total) and provides multi-modal streams including 3D LiDAR, RGB video, IMU, odometry, and joystick commands, collected on two mobile robot platforms (Boston Dynamics Spot and Clearpath Jackal) \cite{scand}. We additionally collect \textit{MMK2-RealNav}, an indoor navigation dataset using the MMK2 robot in a university building, consisting of 18 teleoperated trajectories (1.3km total) with 2 hours of synchronized egocentric RGB observations and planar robot poses. 

\noindent\textbf{Evaluation Metrics}. For trajectory quality, we measure Absolute Trajectory Error (ATE) to quantify global localization accuracy and Relative Pose Error (RPE) to assess local motion consistency between consecutive poses \cite{ATE}. We further evaluate task success with Success Rate (SR), and use Navigation Error (NE) as the distance-to-goal error at the end of an episode. For ACC, we compare predicted observations against ground-truth images using feature-based perceptual similarity, reporting DreamSim \cite{DreamSim}. Finally, to assess overall \emph{visual fidelity} at the distribution level, we compute FID \cite{FID} between the predicted frames and the corresponding ground-truth frames.

\noindent\textbf{Implementation Details}. We initialize MWM from the 1B-parameter variant of a CDiT-XL backbone with a context of 4 frames \cite{NWM}. Teacher-forcing diffusion pretraining is trained with AdamW using a learning rate of $6\times10^{-5}$ and a batch size of 12 on 4$\times$ RTX PRO 6000 (96GB) GPUs with 4 different navigation goals, leading to a final total batch size of 192. ACC post-training is run for 1.5K optimization steps with a learning rate of $2\times10^{-4}$ on a single RTX PRO 6000 (96GB). We use a batch size of 2, where each sample contains an 8-step rollout segment. We found that increasing the number of post-training steps beyond this budget yields negligible additional gains. During self-forcing, we insert LoRA adapters \cite{LORA} into the AdaLN modulation linear layers and train only these adapters, using $r{=}64$, $\alpha{=}16$, and dropout $0.1$.Since the CDiT initialization has already undergone teacher-forcing training on SCAND, we skip Stage~I on SCAND and directly apply self-forcing post-training for experiments on this dataset. For SCAND evaluation, we run CEM for one iteration with 120 sampled action sequences; each candidate is simulated three times and scored by its best outcome to reduce rollout stochasticity. For real-robot evaluation, we use three CEM iterations with 120 samples per iteration, simulating each candidate once. Unless otherwise specified, we use DDIM with 25 denoising steps for NWM and 5 for MWM.

\subsection{Main Results}

\noindent\textbf{ACC}. As shown in Tab.~\ref{tab:lpips_dreamsim_rollout}, MWM achieves lower LPIPS and DreamSim across all rollout horizons compared to NWM, indicating better action-conditioned observation consistency with ground-truth trajectories on SCAND. Notably, MWM under DDIM~5 (which accelerates diffusion inference by reducing the denoising steps from the NWM default of 250 steps to 5 steps using DDIM) not only surpasses NWM with the same accelerated setting, but also outperforms the much slower DDIM setting of NWM (DDIM~25), demonstrating that our post-training enables diffusion inference with at least an 80\% reduction in denoising steps while improving multi-step rollout quality.

\begin{table}[t]
\centering
\caption{Distribution-level visual fidelity on SCAND.}
\label{tab:fid_rollout}
\resizebox{\linewidth}{!}{
\begin{tabular}{l|ccccc}
\toprule
Model & \multicolumn{5}{c}{FID$\downarrow$} \\
\cmidrule(lr){2-6}
 & 1s & 2s & 4s & 8s & 16s \\
\midrule
NWM (DDIM 25) \cite{NWM} & 96.68 & 91.57 & 90.37 & 91.29 & 93.63 \\
NWM (DDIM 5) \cite{NWM}  & 167.43 & 153.87 & 148.70 & 158.13 & 168.28 \\
MWM (ours, DDIM 5) & \textbf{80.97} & \textbf{87.92} & \textbf{85.19} & \textbf{85.80} & \textbf{93.12} \\
\bottomrule
\end{tabular}
}
\end{table}

\begin{table}[t]
\centering
\caption{Inference efficiency measured on an NVIDIA RTX PRO 6000 GPU.}
\label{tab:rollout_time}
\begin{tabular}{l|c}
\toprule
Model & Average Rollout Time (s)$\downarrow$ \\
\midrule
NWM (DDIM 25) \cite{NWM} & 9.6 \\
NWM (DDIM 5) \cite{NWM}  & 2.6 \\
MWM (ours, DDIM 5) & \textbf{2.3} \\
\bottomrule
\end{tabular}
\end{table}

\noindent\textbf{Generation Quality and Inference Efficiency}. As shown in Tab.~\ref{tab:fid_rollout} and Tab.~\ref{tab:rollout_time}, MWM not only achieves lower FID than NWM with DDIM~25, but also provides at least a 4$\times$ speedup. As a comparison, NWM with DDIM~5 suffers a substantial drop in generation quality.

\noindent\textbf{Navigation Performance}. As shown in Tab.~\ref{tab:scand_planning}, MWM achieves the lowest ATE and RPE on SCAND under CEM-based goal-image planning, outperforming all compared baselines and reaching state-of-the-art navigation performance.

\noindent\textbf{Qualitative Results}. Fig.~\ref{fig:benchmark_vis} shows qualitative results on SCAND. The predicted frames exhibit good action-conditioned consistency with the ground-truth frames, remaining temporally coherent and aligned with the executed actions across rollout steps.

\begin{table}[H]
\centering
\caption{Goal-image navigation performance on SCAND with CEM planning.}
\label{tab:scand_planning}
\begin{tabular}{l|cc}
\toprule
Model & ATE$\downarrow$ & RPE$\downarrow$ \\
\midrule
Forward & 2.97 & 0.62 \\
GNM \cite{GNM} & 2.12 & 0.61 \\
NoMaD \cite{NoMaD} & 2.24 & 0.49 \\
NWM + NoMaD ($\times$16) \cite{NWM} & 2.18 & 0.48 \\
NWM + NoMaD ($\times$32) \cite{NWM} & 2.19 & 0.47 \\
NWM \cite{NWM} & 1.28 & 0.33 \\
MWM (ours) & \textbf{1.14} & \textbf{0.302} \\
\bottomrule
\end{tabular}
\end{table}

\begin{figure*}[t]
  \centering
  \includegraphics[width=\textwidth,trim=5mm 45mm 10mm 35mm,clip]{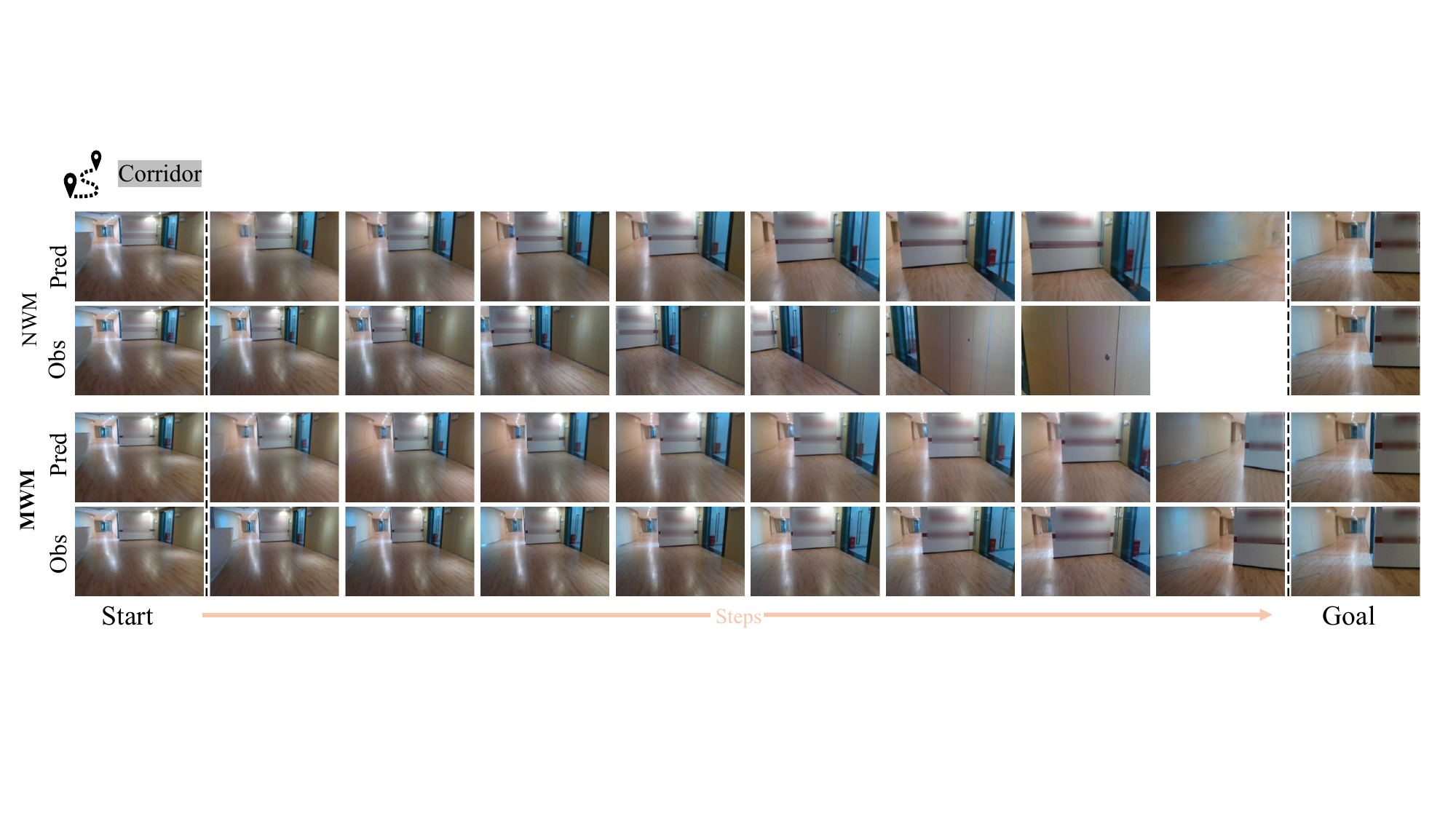}

  \caption{\textbf{Qualitative real-world evaluation}. MWM generates planned rollouts that align better with real observations than NWM, indicating reduced multi-step error accumulation and improved planning quality. Blank frames in the figure indicate cases where the robot was forcibly emergency-stopped due to imminent collision.}
  \label{fig:qual_real}
\end{figure*}

\subsection{Ablation Studies}

We conduct ablation studies to isolate the effects of (i) the ACC loss design, (ii) the two-stage training paradigm, and (iii) ICSD, with results summarized in Tab.~\ref{tab:acc_loss_scand}--\ref{tab:icsd_scand}.

\noindent\textbf{Effect of ACC loss design.} Tab.~\ref{tab:acc_loss_scand} compares different loss choices for ACC on SCAND. Overall, the perceptual LPIPS-based ACC consistently performs best across all three metrics, while pixel-wise objectives (L1/L2) are worse, with L2 being the worst. This suggests that enforcing action-conditioned consistency in a perceptual feature space provides a more suitable learning signal than directly matching pixels.

\noindent\textbf{Effect of training paradigm design.} Since the initial weights of our model have already undergone Stage-I training on SCAND, we study the impact of different training strategies on MMK2-RealNav under this initialization, as shown in Tab.~\ref{tab:paradigm_MMK2-RealNav}. The two-stage pipeline (structure training followed by ACC refinement) achieves the best overall performance. In contrast, training with ACC only yields the worst results, while structure-only training sits in between. These results support our design choice that structure learning establishes a strong predictive prior, and ACC learning further improves action grounding and execution fidelity.

\noindent\textbf{Effect of inference-consistent state.}
Tab.~\ref{tab:icsd_scand} evaluates ICSD on SCAND by varying the context used during training. Using the inference-consistent context $s_{<\tau}^{IC}$ (ICSD) consistently performs better than using $\hat{s}_{<\tau}^{(0)}$ as context. These results suggest that without ICSD, the consistency supervision in ACC post-training is \emph{undermined} by mismatched intermediate states and the step-skipping gap between training and inference.

\begin{table}[h]
\caption{Ablation of ACC loss on SCAND.}
\label{tab:acc_loss_scand}
\centering
\small
\setlength{\tabcolsep}{3pt}
\renewcommand{\arraystretch}{0.95} %
\resizebox{\linewidth}{!}{
\begin{tabular}{l|ccc}
\toprule
Loss & FID (16s)$\downarrow$ & DreamSim (16s)$\downarrow$ & LPIPS (16s)$\downarrow$\\
\midrule
L2 loss            & 352.62 & 0.769 & 0.883\\
L1 loss            & 341.77 & 0.750 & 0.857\\
LPIPS loss (ACC)   & \textbf{93.12} & \textbf{0.337} & \textbf{0.495}\\
\bottomrule
\end{tabular}
}
\end{table}

\begin{table}[h]
\caption{Ablation of training paradigm on MMK2-RealNav.}
\label{tab:paradigm_MMK2-RealNav}
\centering
\small
\setlength{\tabcolsep}{1pt}
\renewcommand{\arraystretch}{0.85} %
\resizebox{\linewidth}{!}{
\begin{tabular}{l|ccc}
\toprule
Training paradigm & FID (16s)$\downarrow$ & DreamSim (16s)$\downarrow$ & LPIPS (16s)$\downarrow$\\
\midrule
Only structure training            & 119.00 & 0.480 & 0.715\\
Only ACC training                  & 196.72 & 0.657 & 0.729\\
Structure training + ACC training  & \textbf{98.17} & \textbf{0.382} & \textbf{0.639}\\
\bottomrule
\end{tabular}
}
\end{table}

\begin{table}[h]
\caption{Ablation of ICSD on SCAND.}
\label{tab:icsd_scand}
\centering
\small
\setlength{\tabcolsep}{3pt}
\renewcommand{\arraystretch}{0.95} %
\resizebox{\linewidth}{!}{
\begin{tabular}{l|ccc}
\toprule
Rollout context in ACC training & FID (16s)$\downarrow$ & DreamSim (16s)$\downarrow$ & LPIPS (16s)$\downarrow$\\
\midrule
$\hat{s}_{<\tau}^{(0)}$            & 109.75 & 0.400 & 0.569\\
$s_{<\tau}^{IC}$ (ICSD)  & \textbf{93.12} & \textbf{0.337} & \textbf{0.495}\\
\bottomrule
\end{tabular}
}
\end{table}

\begin{figure}[h]
    \centering
    \includegraphics[width=\linewidth,trim=0mm 0mm 0mm 0mm,clip]{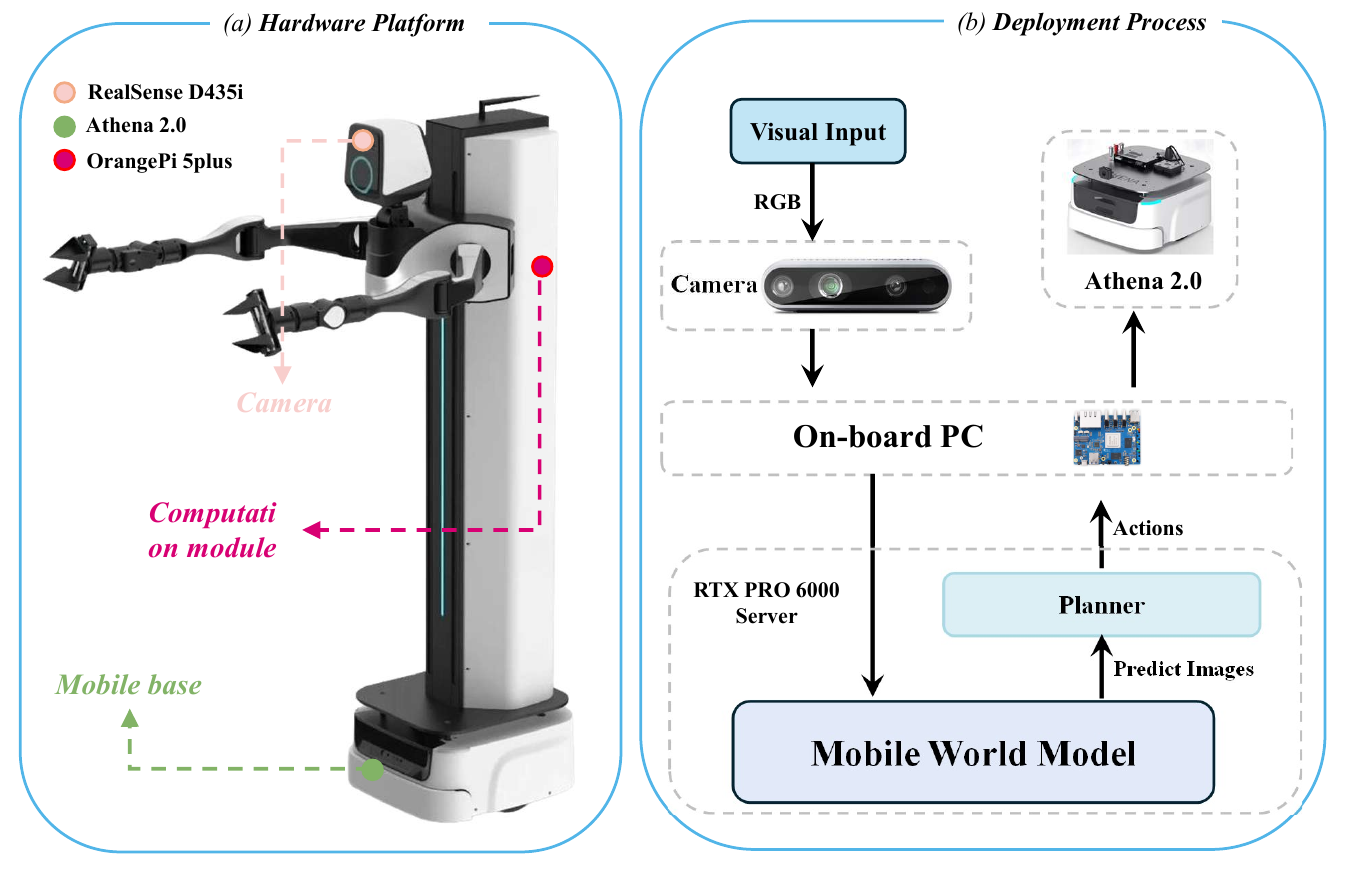}
    \caption{Real-world deployment setup on the AIRBOT Mobile Manipulation Kit 2 (MMK2). 
    (a) Hardware platform.
    (b) Deployment process.}
    \label{fig:mmk2_setupv2}
\end{figure}

\subsection{RealWorld Evaluation}

\subsubsection{Robot Setup}
As shown in Fig.~\ref{fig:mmk2_setupv2}, we use the AIRBOT Mobile Manipulation Kit 2 (MMK2)
mobile manipulation platform for real-world evaluation.
MMK2 is a dual-arm mobile system equipped with multiple
onboard sensors (e.g., vision sensors and LiDAR) and is
designed for mobile operation and data acquisition in indoor
environments. In our experiments, the navigation policies operate on
\emph{RGB-only} observations, without using additional
sensing modalities such as LiDAR or depth. We lower the robot head
to the spine midpoint to obtain a wider and more stable field
of view, and record all RGB observations using the head-mounted camera. All RGB
observations are transmitted to a remote RTX PRO 6000
server for inference, and the returned actions are executed to
control the robot base in a open loop during navigation.

\subsubsection{Scene setup and task types}
We evaluate real-world navigation in an indoor setting in a university building. To cover diverse geometric layouts and visual appearances, we define four representative goal locations: cabinet, window, pillar, and corridor. For each goal, the robot is tasked with goal-image navigation from different starting viewpoints within the same floor, which allows us to assess robustness to viewpoint changes and error accumulation under real-world sensing and actuation noise.
For each episode, we first manually move the robot to the goal location to capture the goal image, then move it to the start location and capture four RGB frames as the initial context.

\begin{table}[h]
\centering
\caption{Quantitative real-world goal-image navigation results aggregated over four targets.}
\label{tab:realworld_nav}
\begin{tabular}{l|cc}
\toprule
Model & SR$\uparrow$ & NE$\downarrow$ \\ 
\midrule
NoMaD \cite{NoMaD} & 0.08 & 2.88\\
NWM \cite{NWM} & 0.20 & 3.18\\
MWM (ours) & \textbf{0.30} & \textbf{2.16}\\
\bottomrule
\end{tabular}
\end{table}

\subsubsection{Quantitative real-world evaluation}
As shown in Tab.~\ref{tab:realworld_nav}, MWM outperforms previous models in real-world goal-image navigation. For fairness, all three methods are evaluated in an open-loop setting, i.e., each method receives only the initial observation frame and performs one-shot planning to the goal. When aggregating results over all four targets, MWM achieves a higher SR and a lower NE, indicating more reliable execution under real-world sensing and actuation noise.

\subsubsection{Qualitative real-world evaluation}
As shown in Fig.~\ref{fig:qual_real}, MWM produces planned rollouts whose best trajectories are more consistent with the robot’s real observations than those generated by NWM. The improved alignment between imagined and observed visual sequences indicates reduced error accumulation during rollout, which directly benefits goal-directed planning and execution. These qualitative results support the effectiveness of our two-stage training framework and help explain the stronger real-world navigation performance achieved by MWM. Fig.~\ref{fig:real} shows the navigation results from a third-person perspective.

\section{CONCLUSIONS}

In this work, we show that combining a two-stage training paradigm with consistency-oriented few-step distillation allows MWM to improve both prediction quality and the practical utility of imagined rollouts for downstream planning. Results on benchmark and real-world tasks further suggest that MWM can serve as an effective interface between visual prediction and navigation control. At the same time, our current system still performs open-loop, one-shot planning: an action sequence is optimized from the current observation and then executed without online replanning from newly observed feedback. For the future work, we aim to extend MWM to real-time closed-loop navigation, where the model continuously updates its imagined rollouts and replans as the robot interacts with dynamic and uncertain environments.

\section*{ACKNOWLEDGMENT}
The authors used ChatGPT (OpenAI) to assist with drafting, revising, and language refinement throughout the main text of this manuscript, including the Introduction, Related Work, Method, Experiments, and Conclusion. The AI-assisted text was reviewed, edited, and verified by the authors.

\bibliographystyle{IEEEtran}
\bibliography{refs}

\end{document}